\pdfoutput=1

\documentclass[11pt]{article}

\usepackage[preprint]{acl}  

\usepackage{times}
\usepackage{latexsym}
\usepackage[T1]{fontenc}

\usepackage[utf8]{inputenc}

\usepackage{microtype}

\usepackage{inconsolata}

\usepackage{graphicx}

\usepackage{graphicx} 
\usepackage{xparse}
\usepackage{url}
\usepackage[normalem]{ulem}
\usepackage{appendix}
\usepackage{color}

\newcommand{\orange}[1]{\textcolor{orange}{#1}}

\newcommand{\rem}[1]{\relax}

\usepackage{hyperref}
\usepackage{xcolor}
\hypersetup{
colorlinks,
linkcolor={red!50!black},
citecolor={blue!50!black},
urlcolor={blue!80!black}
}
\usepackage{booktabs}
\usepackage{subcaption}
\usepackage{ifthen} 
\usepackage{tcolorbox} 
\usepackage{tabularx}

\title{Math Natural Language Inference:\\
this should be easy!}

\author{Valeria de Paiva, Qiyue Gao, Hai Hu, \\
\textbf{Pavel Kovalev,
Yikang Liu, 
Lawrence S.~Moss, Zhiheng Qian}}

\newcommand{\Addresses}{\noindent
Valeria de Paiva, Topos Institute, \texttt{valeria@topos.institute} \\
Qiyue Gao, University of California, San Diego, \texttt{q3gao@ucsd.edu} \\
Hai Hu, Shanghai Jiao Tong University, \texttt{hu.hai@outlook.com} \\
Pavel Kovalev, Carnegie Mellon University, \texttt{pkovalev@andrew.cmu.edu}\\
Yikang Liu,  Shanghai Jiao Tong University, \texttt{yikangliu@sjtu.edu.cn} \\
Lawrence S.~Moss, Indiana University, \texttt{lmoss@iu.edu}\\
Zhiheng Qian, Shanghai Jiao Tong University, \texttt{n1vnhil@sjtu.edu.cn}
}

\rem{
\author{
 \textbf{Valeria de Paiva\footnote{valeria@topos.institute}},
  \textbf{Qiyue Gao\footnote{2}},
  \textbf{Hai Hu\footnote{3}},\\
      \textbf{Yikang Liu\footnote{yikangliu@sjtu.edu.cn}},
  \textbf{Pavel Kovalev\footnote{pkovalev@andrew.cmu.edu}},
  \textbf{Lawrence S.~Moss\footnote{lmoss@iu.edu}},
    \textbf{Zhiheng Qian\footnote{n1vnhil@sjtu.edu.cn}}
}
}

\date{August 2025}

\NewDocumentCommand{\bert}{ mO{} }{\textcolor{blue}{\textsuperscript{\textit{Bert}}\textsf{\textbf{\small[#1]}}}}

\begin{document}

\maketitle

\begin{abstract}
We ask whether contemporary LLMs are able to perform natural language inference (NLI) tasks on mathematical texts.  We call this the Math NLI problem.  We construct a corpus of Math NLI pairs whose premises are from extant mathematical text and whose hypotheses and gold labels were provided by people with experience in both research-level mathematics and also in the NLI field.   We also investigate the quality of corpora using the same premises but whose hypotheses are provided by LLMs themselves.  We not only investigate the performance but also the inter-group consistency of the diverse group of LLMs.   We have both positive and negative findings.   Among our positive findings: in some settings, using a majority vote of LLMs is approximately equivalent to using human-labeled data in the Math NLI area.   On the negative side: LLMs still struggle with mathematical language.  They occasionally fail at even basic inferences.  Current models are not as prone to hypothesis-only ``inference'' in our data the way the previous generation had been.    In addition to our findings, we also provide our corpora as data to support future work on Math NLI.
\end{abstract}

\section{Introduction}

We study natural language inference 
(NLI) tasks in the area of research-level mathematics.
One might think that LLMs would do extremely well on this task. 
After all, what counts as an entailment or contradiction in everyday-language texts is often taken as a complicated version of what happens with mathematics.  So we 
might expect purported mathematical inferences to be \emph{easier} to 
evaluate than those in everyday language.  And unlike language in the wild,
the domain of mathematics is fairly well-defined. 
Facts, definitions, and logical reasoning play a large role in mathematical writing.
Sentences ought to be precise and unambiguous.

However, there are complications with mathematical text from the start.  The vocabulary may be unfamiliar to a generic audience: mathematical parlance can use daily words with new,  unfamiliar meanings, e.g. `ring', `field', or even `folklore'. On top of this, the use of visual elements such as symbols, equations, and diagrams, almost
changes the very language of the text from plain text to a richer, multimodal language. 
%
The field lacks open-source resources such as dictionaries and glossaries for mathematical concepts.
It is much harder to find a ``person on the street'' annotator of mathematics than of
more common forms of text.  

When confronted with the incredible solutions to mathematical-like problems that deep learning systems can offer nowadays (e.g., AlphaGeometry~\cite{Trinh2024}), it is difficult to believe that these systems cannot understand the basics of causality or of propositional reasoning used throughout mathematics. Nonetheless, when tested on these basics,
the LLMs-based systems still make very surprising (to humans) mistakes. Further, the fact that LLMs do not have a notion of self-consistency has been documented in many recent papers 
\cite{sedova-etal-2024,kıcıman2024,xu2024llmsreally}. 
But mathematics, as usually practiced, needs self-consistency.
In a sense, it seems that sometimes the deep learning systems deserve an A+ in advanced problem solving but a B in the basics.

%
%
For all of these reasons, we could conclude, 
perhaps surprisingly, that the NLI task is not much easier when using LLMs to deal with mathematical text after all.  In this paper, we shall see how precisely correct math NLI using LLMs can be. We decided to experiment and build a corpus of NLI inference pairs, comparing the output of several LLMs on mathematical text.

\subsection{Research questions}

\begin{table*}[t]
\centering
\small
\begin{tabular}{p{7cm}p{7cm}p{1cm}}
\toprule
\textbf{P (Premise)} & \textbf{H (Hypothesis)} & \textbf{Label} \\
\midrule
\textit{A notion of central importance in categorical topology is that of topological functor.} & Topological functor is a notion of categorical topology. & E \\
\addlinespace
\textit{The problem of relating a factorization system to a pointed endofunctor is considered.} & The problem of relating a factorization system to a pointed endofunctor is not discussed. & C \\
\addlinespace
\textit{A notion of central importance in categorical topology is that of topological functor.} & There are many notions of central importance in categorical topology. & N \\
\bottomrule
\end{tabular}
\caption{Examples in human-created seed Math NLI  corpus. \label{tab:examples:human:corpus}}
\end{table*}

Our big question:
Can LLMs be reliable constructors and annotators of math NLI corpora?
We address this by asking and answering some secondary questions:
(a) How well do LLMs perform on a math NLI corpus annotated by mathematicians?
(b) Are there common features to the errors which they make?
(c) How good is a math NLI corpus annotated entirely by LLMs?
(d) Are LLMs more unanimous on human-written corpora or on corpora generated by LLMs themselves?
    
\subsection{Goal, plan and structure of the paper}

The ``deliverables'' of this paper are two corpora for Math NLI: one written by humans and the other by GPT.   These are not benchmarks. But we believe that they will help others who work on this topic.

Equally important, this paper details what we have learned about Math NLI from several years of work, including work that did not turn out as well as we had hoped.   Overall, our goal is to make some points about Math NLI which we believe have not been made elsewhere, based on data and examples which we have collected.  The plan of the paper is to tell the story of this work.


\section{Math NLI seed corpus
}

\subsection{Creation of a seed set of pairs}

Our first experiment 
used a corpus of abstracts of articles in the journal \textit{Theory and Applications of Categories} (TAC) developed in \cite{collard2022}\footnote{Available at \url{https://github.com/ToposInstitute/tac-corpus}.}.  This corpus has some 3K sentences, but 432 were singled out as `Goldilocks-like sentences': not too short, not too long, and with  little or no \LaTeX\ markup.  Then  we chose about 30 of these sentences, and for each sentence $S$ in this set, three of 
our team members were asked to write a sentence entailed by $S$, a sentence contradicting $S$, and a sentence neutral with respect to $S$. 
(So we had the ``gold labels'' by construction.  But as we found repeatedly, getting consistent data from humans is difficult, even about mathematical texts.)
The team members were told to produce
grammatical sentences that did not depend on factual knowledge about the mathematics in the original TAC sentence and that tried to introduce as few new facts as possible. It is impossible to do this perfectly, but the team members strove to do so.   
As a result we had $3 \times 3 \times 31 = 279$ pairs, equally divided with $E$, $C$, and $N$ labels.

We aimed to fulfill the following conditions as much as possible:
\begin{enumerate}
\item Inferences should be uncontroversial. 
We want inferences which most mathematicians would take to be ``immediate.''
\item We treat mathematical concepts as black boxes. (Inference should depend  as little as possible on the background mathematical knowledge of the assessor.)
\item  We avoid ``dangling references", pronouns (it, they) or demonstratives (this, that, here, there) without clear antecedents.  In general, we 
tried to avoid all of the problematic issues in natural language semantics.
\end{enumerate}

Table~\ref{tab:examples:human:corpus} shows some examples of human-created hypotheses and their labels.

\begin{table}[t]
\centering
\small
\begin{tabular}{ll}\toprule
     Abbr. & Model \\\midrule  
     GPT4 & GPT4\footnote{\url{https://openai.com/}} \\
     L2 & Llama-2\\ 
     L3 & Llama-3.0\\ 
     C3 & Claude-3\\\ 
     Mistral & Mistral-large \\\hline  
     L3.1 & Llama-3.1-70B-Instruct\\ 
     Q2 & Qwen2-72B-Instruct \\
     Mixtral & Mixtral-8x22B-Instruct-v0.1 \\ 
     DS & deepseek-llm-67b-chat\\
     Ge2 & gemma-2-27b-it 
     
\end{tabular}
\caption{LLMs used in Exp.~1. Top: Group 1: five initial LLMs; Bottom: Group 2: five later LLMs. }
\label{tab:10:LLMs}
\end{table}

Having constructed our seed set of 279 pairs
we used a collection of LLMs to evaluate it, as shown in Table~\ref{tab:10:LLMs}. 
This led to the realization that not only did human creators disagree with each other, also the rate of unanimity between machines was not very stable. In particular, we discovered some 20 pairs with contradictory evaluations between machines and humans. We called these the \textit{red pairs}, as they deserved further attention. 
We explain our process of evaluation, the LLMs used, and our set up in the next section, but we discuss briefly the red pairs now.

\subsection{Red Pairs} 

Our three mathematically-trained group memebers tried to analyze the kinds of mistakes LLMs were making in these pairs. 
We discovered a few patterns of problematic or flawed reasoning:

\paragraph{Ignored context.}
 Sometimes a specific context was mentioned, for instance
\begin{itemize}
 \item 
    \textit{P: In the nilpotent case, this nerve is known to be a Kan complex.}\\
    {H: This nerve is not known to be a Kan complex.}  
\end{itemize}
but the LLMs discarded the specific context (\textit{the nilpotent case}) and compared the matrix sentences -- in the example above this leads to a contradiction --  instead of a neutral label. This is similar to the problems with modal and counterfactual reasoning discussed in \cite{holliday2024}.

\paragraph{Vague quantifiers.}  We also have problems with vague predicates like \textit{numerous, few, many}, where humans could also disagree amongst themselves: one example from the `red pairs' set is 
\begin{itemize}
 \item 
    \textit{P: We worked through numerous examples to demonstrate the power of these notions.} \\ 
    {H: We worked through two examples to demonstrate the power of these notions.}  
\end{itemize}
The mathematicians agreed that \textit{numerous examples} should entail \textit{two examples}, but LLMs did not.

\paragraph{Lexical ambiguity.} There is lexical ambiguity, for example,  with the verb ``resemble" which might mean 
``is almost equal" (for some humans) or  ``it looks similar to something else, but it is
not the same as" -- a reason why we might have humans saying both contradiction or entailment in the example:
\begin{itemize}
 \item 
    \textit{P: The axioms resemble those for monoidal Abelian categories.} \\
    {H: The axioms are the ones of monoidal Abelian categories.}  
\end{itemize}
Note that the ambiguity which we call ``lexical'' here might also be called ``pragmatic'' because the issue is whether the use of ``resemble'' here carries the Gricean implicature that if an object $A$ resembles an object $B$, then $A$ is not, strictly speaking, $B$ at all.

\paragraph{Naming of math entities.} There is a problem with naming mathematical entities, e.g. ``group B" vs. ``group C" if this is only used as a generic name, as an $\alpha$-variant, then the difference between B and C doesn't matter. But many times we are talking about different groups.

\paragraph{Unknown math concepts.} Sometimes one really must know the concepts involved. For example, for the pair 
\begin{itemize}
 \item 
    \textit{P: This paper proposes a recursive definition of V-n-categories and their morphisms.}\\
    {H: This paper proposes a definition of V-categories.}  
\end{itemize}
if we know that `V-n-categories' are `V-categories', then we can decide on entailment. But how do we know that? The mathematician is at liberty to create concepts and name them in strange ways. For instance a ``skew monoidal category'' is not a ``monoidal category'', only an `almost' monoidal category.

\vspace{1em}
\section{
Evaluating LLMs on the seed corpus}
In our first experiment, we harness LLMs to evaluate the seed corpus.

\subsection{Method}

The seed corpus was originally judged by five LLMs, the top ones in Table~\ref{tab:10:LLMs}. 
We used the prompt shown in Appendix~\ref{section-seed-corpus-prompt}.
When 4 or 5 LLMs disagreed with the human annotation, we discussed the pair again, throwing it out if it was considered ``controversial'' by the mathematicians in our group.

We use API services from \texttt{together.ai} to query the LLMs,  using a script to
extract E/C/N judgments from each model's
explanation. The algorithm used is simple: it counts the occurrences of a few keywords in the first sentence without semantical analysis. 
(It works well if the model gives the answer directly.)
However, this algorithm can fail.  For example, when the model does not follow the instructions strictly we may end up with a pair that is neither E nor C nor N, and as usual in NLI we take N as a catch-all for ``not E and not C.''

\subsection{Results}

\begin{table*}[t]
\centering
\begin{tabular}{llllllllllll}\toprule
      &    & GPT4 & L2   & L3   & C3   & Mistral & L3.1 & Q2   & Mixtral & DS   & Ge2  \\ \midrule
     & p  & 82.9 & 90.5 & 70.9 & 91.8 & 79.8    & 88.8 & 87.3 & 75.4    & 92.7 & 85.1 \\
C      & r  & 98.9 & 61.3 & 96.8 & 95.7 & 97.8    & 93.5 & 95.7 & 98.9    & 81.7 & 92.5 \\
      & f1 & 90.2 & 73.1 & 81.8 & 93.7 & 87.9    & 91.1 & 91.3 & 85.6    & 86.9 & 88.7 \\\midrule
     & p  & 90.1 & 73.9 & 85.4 & 93.5 & 89.8    & 83.8 & 80.8 & 86.3    & 82.0 & 82.8 \\
E      & r  & 97.8 & 88.2 & 94.6 & 92.5 & 84.9    & 89.2 & 90.3 & 88.2    & 78.5 & 82.8 \\
      & f1 & 93.8 & 80.4 & 89.8 & 93.0 & 87.3    & 86.5 & 85.3 & 87.2    & 80.2 & 82.8 \\\midrule
     & p  & 95.5 & 56.2 & 91.8 & 87.8 & 81.8    & 81.7 & 84.7 & 85.5    & 67.6 & 75.3 \\
N      & r  & 68.8 & 63.4 & 48.4 & 84.9 & 67.7    & 72.0 & 65.6 & 57.0    & 78.5 & 68.8 \\
      & f1 & 80.0 & 59.6 & 63.4 & 86.3 & 74.1    & 76.6 & 73.9 & 68.4    & 72.6 & 71.9 \\\midrule
acc   &    & 88.5 & 71.0 & 79.9 & \textbf{91.0} & 83.5    & 84.9 & 83.9 & 81.4    & 79.6 & 81.4 \\\midrule
      & p  & 89.5 & 73.5 & 82.7 & 91.0 & 83.8    & 84.8 & 84.2 & 82.4    & 80.8 & 81.1 \\
avg      & r  & 88.5 & 71.0 & 79.9 & 91.0 & 83.5    & 84.9 & 83.9 & 81.4    & 79.6 & 81.4 \\
      & f1 & 88.0 & 71.0 & 78.3 & 91.0 & 83.1    & 84.7 & 83.5 & 80.4    & 79.9 & 81.1\\\bottomrule
\end{tabular}

    \caption{Results of 10 LLMs on the seed MathNLI corpus. 
  On the left we list the initial models, and starting with L3.1  the recent ones. }
    \label{tab:res:5:new:LLMs:seed}
\end{table*}




Performance of 10 LLMs on the seed MathNLI corpus is shown in Table~\ref{tab:res:5:new:LLMs:seed}, with their confusion matrices shown in Table~\ref{tab:seed:corpus:confusion:matrices}.

Table~\ref{tab:res:5:new:LLMs:seed} presents the precision, recall, f1-score and accuracy for 10 LLMs. The overall accuracy is medium to high, ranging from 71\% to 91\%, suggesting that in general, the LLMs we tested can perform category-theory-related mathematical inference to a certain degree. 
We note that the first group of LLMs (to the left of the table) are not particularly better than the second group (on the right).
This might reflect the fact that the first group were closed-source,
while Group 2's models were open-source.  
The first group has two closed source models: Claude 3 and GPT4; the others are open source.
In particular, Claude 3 seems to still be better than the open-source LLMs, but perhaps more runs are necessary to confirm this.

\begin{table*}[ht]
\centering
\begin{subtable}[t]{0.25\textwidth}
\centering
\caption{GPT4}
\footnotesize
\begin{tabular}{lrrr}
\toprule
Gold$\downarrow$ & C & E & N \\
\midrule
C& 98.9 & .0 & 1.1 \\
E& .0 & 97.8 & 2.2 \\
N& 20.4 & 10.8 & 68.8 \\
\bottomrule
\end{tabular}
\end{subtable}
\hfill
\begin{subtable}[t]{0.16\textwidth}
\centering
\footnotesize
\caption{Llama2}
\begin{tabular}{rrr}
\toprule
C & E & N \\
\midrule
 61.3 & 1.1 & 37.6 \\
 .0 & 88.2 & 11.8 \\
 6.5 & 30.1 & 63.4 \\
\bottomrule
\end{tabular}
\end{subtable}
\hfill
\begin{subtable}[t]{0.16\textwidth}
\centering
\footnotesize
\caption{Llama3}
\begin{tabular}{rrr}
\toprule
 C & E & N \\
\midrule
96.8 & .0 & 3.2 \\
4.3 & 94.6 & 1.1 \\
35.5 & 16.1 & 48.4 \\
\bottomrule
\end{tabular}
\end{subtable}
\hfill
\begin{subtable}[t]{0.16\textwidth}
\centering
\footnotesize
\caption{Claude3}
\begin{tabular}{rrr}
\toprule
  C & E & N \\
\midrule
 95.7 & .0 & 4.3 \\
 .0 & 92.5 & 7.5 \\
 8.6 & 6.5 & 84.9 \\
\bottomrule
\end{tabular}
\end{subtable}
\hfill
\begin{subtable}[t]{0.16\textwidth}
\centering
\footnotesize
\caption{Mistral}
\begin{tabular}{rrr}
\toprule
  C & E & N \\
\midrule
 97.8 & .0 & 2.2 \\
 2.2 & 84.9 & 12.9 \\
 22.6 & 9.7 & 67.7 \\
\bottomrule
\end{tabular}
\end{subtable}

\vspace{1em}

\begin{subtable}[t]{0.20\textwidth}
\centering
\footnotesize
\caption{Llama3}
\begin{tabular}{lrrr}
\toprule
 & C & E & N \\
\midrule
 C&93.5 & .0 & 6.5 \\
 E&1.1 & 89.2 & 9.7 \\
 N&10.8 & 17.2 & 72.0 \\
\bottomrule
\end{tabular}
\end{subtable}
\hfill
\begin{subtable}[t]{0.16\textwidth}
\centering
\footnotesize
\caption{Qwen2}
\begin{tabular}{rrr}
\toprule
  C & E & N \\
\midrule
 95.7 & .0 & 4.3 \\
 2.2 & 90.3 & 7.5 \\
 12.0 & 21.7 & 66.3 \\
\bottomrule
\end{tabular}
\end{subtable}
\hfill
\begin{subtable}[t]{0.16\textwidth}
\centering
\footnotesize
\caption{Mixtral}
\begin{tabular}{rrr}
\toprule
 C & E & N \\
\midrule
 98.9 & .0 & 1.1 \\
 3.2 & 88.2 & 8.6 \\
 29.0 & 14.0 & 57.0 \\
\bottomrule
\end{tabular}
\end{subtable}
\hfill
\begin{subtable}[t]{0.16\textwidth}
\centering
\footnotesize
\caption{DeepSeek}
\begin{tabular}{rrr}
\toprule
 C & E & N \\
\midrule
 81.7 & .0 & 18.3 \\
 2.2 & 78.5 & 19.4 \\
 4.3 & 17.2 & 78.5 \\
\bottomrule
\end{tabular}
\end{subtable}
\hfill
\begin{subtable}[t]{0.16\textwidth}
\centering
\footnotesize
\caption{Gemma2}
\begin{tabular}{rrr}
\toprule
C & E & N \\
\midrule
 92.5 & 1.1 & 6.5 \\
 1.1 & 82.8 & 16.1 \\
 15.1 & 16.1 & 68.8 \\
\bottomrule
\end{tabular}
\end{subtable}
\caption{Confusion Matrices Comparison for 10 LLMs on the seed MathNLI corpus. 
\label{tab:seed:corpus:confusion:matrices}}
\end{table*}

A main message from Table~\ref{tab:seed:corpus:confusion:matrices} is that most models struggle with \emph{neutral pairs}, mistakenly categorizing them either as entailment pairs or contradictory pairs. 
For instance, Llama-3 is particularly bad in that it labels as many as 35\%  of neutral pairs as contradictions; only 48\% of the neutral pairs are correctly classified. Claude3 is the best in labeling N pairs, with an accuracy of 84.9\% for them.  
On the contrary, most models perform very well on C and E pairs. GPT4, Llama3 and Qwen2 correctly labeled more than 90\% of the C and E pairs. 
In fact, C pairs are the easiest for all models, except Llama2, with most models achieving accuracy greater than 90\%. 
Furthermore, models seldom confuse C and E pairs. For eight out of the ten LLMs, C pairs are never categorized as E pairs. 

Only one pair in one model (Gemma2) is classified as  C by the machines  and   E by humans:
\begin{itemize}
    \item \textit{P: Both of them generalise the concept of algebra on a monad T.}\\
    {H: The concept of algebra on a monad T is more special than both of them.}   
\end{itemize}
Note that this pair does not satisfy our criteria of explicit references only. The pair is fairly controversial, as well. All  LLMs label it as contradictory, but mathematicians tend to think that generalizing and specializing are antonyms. So whatever "both of them" are, if they are a generalization of the concept of algebra of a monad (as claimed by the premise) then "algebra of a monad" will more specialized than them.


Concerning the Group 1 models: out of 279 samples, there is at least one model that agrees with the human annotator in 271 samples.
Hence, there are 8 pairs where none of the 5 initial models agrees with the human label. These  eight pairs are recalled in  Appendix A. The examples are telling as they point out patterns of reasoning that might be difficult for humans as well. For instance:

\begin{itemize}
 \item 
    \textit{P: Using these ideas, we also prove that magnetic monopoles form an abelian group.} {H: Using these ideas, we also prove that monopoles form an abelian group.}  
\end{itemize}

\noindent Clearly a mathematician would gather that `magnetic poles' form an abelian group, but nothing has been said about non-magnetic poles. So neutral is much more reasonable than `entailment'. (More on this is in the appendix A).


Table~\ref{tab:exp1:res:unanimous} discusses unanimity between LLMs.
As before we consider two groups of models.
Our initial LLMs are unanimous in 163 of the pairs ($58.4\%$).
Of these 163, in 155 of the cases, the models' agreed-upon label matches the human annotations.  And in 271 of the 290 pairs (including ones where the models were not unanimous), at least one model agreed with the
human label.  This explains the upper row of the table, and the lower row is similar.


Notice that for the 
more recent LLMs, unanimity goes up from  $58.4\%$ to $68.1\%$.
We do not have a good explanation of this.


\begin{table*}[t]
    \centering
    \begin{tabular}{lccc}\toprule
         & unanimous & some agree w/ human &  agrees w/ a human  \\ \midrule
    models in group 1 & 163 (=58.4\%) & 271 (=97.1\%) & 155 (=55.6\%) \\
    models in group 2 & 190 (=68.1\%) & 266 (=95.3\%) & 178 (=63.8\%) \\ \midrule
    \end{tabular}
    \caption{Agreement by LLMs on the seed corpus.}
    \label{tab:exp1:res:unanimous}
\end{table*}

\vspace{2em}


\section{
Using LLMs to generate  a MathNLI corpus}


\subsection{Generation using GPT4} Our second experiment asked GPT-4 to generate Entailment, Contradiction, and Neutral hypotheses from the Goldilocks sentences in the TAC corpus, resulting in 1157  pairs.
The prompt we used is shown below:
\begin{tcolorbox}[colframe=gray!75!black, boxsep=1mm, left=2mm, right=2mm, top=1mm, bottom=1mm] 
Generate ``Entailment``, ``Contradiction``, ``Neutral`` hypothesis of a given sentence.\\
Here are some examples:
        \texttt{[example\_script]}
        
        Sentence: \texttt{[context]}
\end{tcolorbox}
GPT-4 was a good generator of pairs, as we shall see below.
But it was not consistent with itself. If it created a pair nominally to be E it could later judge it N or even C.
As we  see in Table 6,
41.4\% of the pairs which
GPT-4 created to be neutral it later claims as entailments.

\subsection{Checking of a subset, using both humans and LLMs}
We chose 89 pairs to conduct manual evaluation and distributed these among the mathematicians of the group. This gave us a set of 89 GPT4-created/human evaluated pairs.
These 89 pairs were also evaluated using GPT4, Llama2, Llama3 and Claude3, in the first instance. 
Our mathematicians agree with each other in 80 of the 89 pairs. 
They agree with 74 (83\%) of the GPT-generated labels. 




\section{
Evaluating LLMs on GPT-generated MathNLI corpus}







Next, we had the 4 models in Group 1 and 5 models in Group 2 label the 89 pairs.
The results are shown in Table \ref{tab:expr3-result}.
The models in group 1 show unanimous agreement in 57 of the pairs (64\%), while the models
in group 2 do so in 65  (73\%). In group 1, for 50 of these 65 pairs (87\%), their unanimous label agrees with human labels; while the agreement for group 2 is 57 pairs (88\%).
Here is our conclusion from this experiment:
If we take the unanimous labels from the group 2 models to simply \emph{be} the gold label, then this label is the same as the human label 88\% of the time.


\begin{table*}[ht]
    \centering
    \begin{tabular}{lccc} \toprule
    & unanimous & agree w/ at least 1 human &  agree w/ all human\\ 
    \midrule
    human annotator & / & / & 80 (= 89.9\%) \\
    GPT generator & /& 74 (= 83.1\%) & 65 (= 73.0\%)\\
    models in group 1 & 57 (= 64.0\%) & 50 (= 56.2\%) & 43 (= 48.3 \%)\\
    models in group 2 & 65 (= 73.0\%) & 57 (= 64.0\%) & 50 (= 56.2\%)\\ \bottomrule
    \end{tabular}
    \caption{Experiment 3 Result: total 89 pairs generated by GPT4}
    \label{tab:expr3-result}
\end{table*}



\begin{table*}[ht]
\centering
\begin{subtable}[t]{0.14\textwidth}
\centering
\caption{GPT4}
\footnotesize
\begin{tabular}{lrrr}
\toprule
Gold$\downarrow$ & C & E & N \\
\midrule
C & 96.7 & .0 & 3.3 \\
E & .0 & 96.7 & 3.3 \\
N & .0 & 41.4 & 58.6 \\
\bottomrule
\end{tabular}
\end{subtable}
\hfill
\begin{subtable}[t]{0.14\textwidth}
\centering
\footnotesize
\caption{Llama2}
\begin{tabular}{lrrr}
\toprule
 C & E & N \\
\midrule
 53.3 & .0 & 46.7 \\
 .0 & 100.0 & .0 \\
 .0 & 75.9 & 24.1 \\
\bottomrule
\end{tabular}
\end{subtable}
\hfill
\begin{subtable}[t]{0.14\textwidth}
\centering
\footnotesize
\caption{Llama3}
\begin{tabular}{lrrr}
\toprule
 C & E & N \\
\midrule
 96.7 & .0 & 3.3 \\
 .0 & 100.0 & .0 \\
 .0 & 51.7 & 48.3 \\
\bottomrule
\end{tabular}
\end{subtable}
\hfill
\begin{subtable}[t]{0.14\textwidth}
\centering
\footnotesize
\caption{Claude3}
\begin{tabular}{lrrr}
\toprule
 C & E & N \\
\midrule
 93.3 & 3.3 & 3.3 \\
 .0 & 93.3 & 6.7 \\
 3.4 & 34.5 & 62.1 \\
\bottomrule
\end{tabular}
\end{subtable}

\vspace{1em}

\begin{subtable}[t]{0.14\textwidth}
\centering
\footnotesize
\caption{Llama3.1}
\begin{tabular}{lrrr}
\toprule
 C & E & N \\
\midrule
 93.3 & .0 & 6.7 \\
 .0 & 100.0 & .0 \\
 3.4 & 55.2 & 41.4 \\
\bottomrule
\end{tabular}
\end{subtable}
\hfill
\begin{subtable}[t]{0.14\textwidth}
\centering
\footnotesize
\caption{Qwen2}
\begin{tabular}{rrr}
\toprule
C & E & N \\
\midrule
 93.3 & .0 & 6.7 \\
 .0 & 100.0 & .0 \\
 3.4 & 34.5 & 62.1 \\
\bottomrule
\end{tabular}
\end{subtable}
\hfill
\begin{subtable}[t]{0.14\textwidth}
\centering
\footnotesize
\caption{Mixtral}
\begin{tabular}{rrr}
\toprule
 C & E & N \\
\midrule
 96.7 & .0 & 3.3 \\
 .0 & 96.7 & 3.3 \\
 3.4 & 31.0 & 65.5 \\
\bottomrule
\end{tabular}
\end{subtable}
\hfill
\begin{subtable}[t]{0.14\textwidth}
\centering
\footnotesize
\caption{Deepseek}
\begin{tabular}{rrr}
\toprule
 C & E & N \\
\midrule
 83.3 & .0 & 16.7 \\
 .0 & 90.0 & 10.0 \\
 .0 & 31.0 & 69.0 \\
\bottomrule
\end{tabular}
\end{subtable}
\hfill
\begin{subtable}[t]{0.14\textwidth}
\centering
\footnotesize
\caption{Gemma2}
\begin{tabular}{rrr}
\toprule
 C & E & N \\
\midrule
 100.0 & .0 & .0 \\
 .0 & 90.0 & 10.0 \\
 .0 & 34.5 & 65.5 \\
\bottomrule
\end{tabular}
\end{subtable}
\caption{Confusion Matrices on GPT-generated Corpus}
\label{tab:confusion:matrices:gpt:gen}
\end{table*}

The evaluation results on the GPT-generated corpus using 
the GPT-generated label as the true label are shown in Table \ref{tab:res:gpt:gen:corpus}, with the confusion matrices presented in Table \ref{tab:confusion:matrices:gpt:gen}. The overall accuracy of LLMs varies between 59.6\% and 86.5\%, which is relatively lower than the accuracy on the seed corpus.

Our analysis reveals that while the E and C pairs generated by GPT show a certain level of consistency relative to our seed pairs, N pairs are frequently misclassified as E. (This finding echoes what we saw
in our previous experiment, but there the pairs were human-generated.) Surprisingly, Llama2 classifies 75.9\% of N pairs as E. 
Among all evaluated models, Mixtral showed the least susceptibility to this issue, maintaining the highest accuracy of 76.0\%. Although its performance on the seed corpus was not outstanding, Mixtral achieved the highest overall accuracy of 86.5\% on the GPT-generated corpus.

Furthermore, it was observed that LLMs tend to identify C pairs within the GPT-generated corpus more accurately than they do within the seed corpus; 8 out of 9 models achieved an F1 score of over 90\%. Notably, Gemma2 successfully detected all the C samples in the GPT-generated corpus.

\begin{table*}[ht]
\begin{tabular}{lllllllllll}\toprule
& & GPT4 & L2 & L3 & C3 & L3.1 & Q2 & Mixtral & DS & Ge2 \\\midrule
 & precision & 100.0 & 100.0 & 100.0 & 96.6 & 96.6 & 96.6 & 96.7 & 100.0 & 100.0 \\
C & recall & 96.7 & 53.3 & 96.7 & 93.3 & 93.3 & 93.3 & 96.7 & 83.3 & 100.0 \\
& f1-score & 98.3 & 69.6 & 98.3 & 94.9 & 94.9 & 94.9 & 96.7 & 90.9 & \textbf{100.0} \\\midrule
 & precision & 70.7 & 57.7 & 66.7 & 71.8 & 65.2 & 75.0 & 76.3 & 75.0 & 73.0 \\
E & recall & 96.7 & 100.0 & 100.0 & 93.3 & 100.0 & 100.0 & 96.7 & 90.0 & 90.0 \\
& f1-score & 81.7 & 73.2 & 80.0 & 81.2 & 78.9 & \textbf{85.7} & 85.3 & 81.8 & 80.6 \\\midrule
 & precision & 89.5 & 33.3 & 93.3 & 85.7 & 85.7 & 90.0 & 90.5 & 71.4 & 86.4 \\
N & recall & 58.6 & 24.1 & 48.3 & 62.1 & 41.4 & 62.1 & 65.5 & 69.0 & 65.5 \\
& f1-score & 70.8 & 28.0 & 63.6 & 72.0 & 55.8 & 73.5 & \textbf{76.0} & 70.2 & 74.5 \\\midrule
acc &  & 84.3 & 59.6 & 82.0 & 83.1 & 78.7 & 85.4 & \textbf{86.5} & 80.9 & 85.4 \\\midrule
 & precision & 86.7 & 64.0 & 86.6 & 84.7 & 82.5 & 87.2 & 87.8 & 82.3 & 86.4 \\
avg & recall & 84.3 & 59.6 & 82.0 & 83.1 & 78.7 & 85.4 & 86.5 & 80.9 & 85.4 \\
& f1-score & 83.8 & 57.2 & 80.8 & 82.8 & 76.8 & 84.8 & \textbf{86.1} & 81.1 & 85.2\\\bottomrule
\end{tabular}
\caption{Results of LLMs on GPT-generated Corpus.}
\label{tab:res:gpt:gen:corpus}
\end{table*}

\section{Sanity checks}
For previous NLI datasets such as SNLI~\citep{snli} and MNLI~\citep{mnli}, studies have shown that some of the systems which detect ``entailment'' were working on artifacts of the datasets, instead of doing any logical reasoning~\citep[e.g.,][]{poliak-etal-2018}.
In this section we provide some sanity checks on our corpus.

\subsection{Does every assertion entail itself?}
We are trying to make reasoning as clear as possible for humans and systems, 
and in some sense as simple as possible.  We thus 
 propose the easiest logical deduction that we expect LLMs (and humans) to make proficiently. 
Taking a cue from propositional logic, this would be that every proposition $A$ entails itself, $A\to A$. To see how much noise LLMs could introduce, we ask them
whether every one of a set of mathematical assertions entailed itself. It was not very surprising to discover that for one of 180 such pairs, one LLM missed this assertion $A\to A$.

\begin{itemize}
 \item 
    \textit{P: We do not include any guiding examples.} {H: We do not include any guiding examples.}  
\end{itemize}
This pair was judged N instead of E.
But the explanation for the reasoning of the LLM is fascinating: 

\begin{quote}
    Explanation: In this case, the premise and the hypothesis are identical statements. Since they both convey the same information, there is no logical inference or contradiction between them. The hypothesis is neither entailed by the premise nor does it contradict the premise. Therefore, the hypothesis is neutral to the premise.
\end{quote}

The first sentence is correct, but immediately following that
the model blunders.

The experiment was repeated with many more pairs, and only 3 in 2312 pairs had the same problem. This is a very small number of errors, but this is arguably the easiest inference one can ever make. It is odd that a system which can make deep, complicated mathematical inferences,  can get a simple inference like `$A$ implies $A$' wrong.

Previous work such as \citet{xu2024llmsreally} tries to catalog the kinds of mistakes that LLMs are known to make. They suggest that ``to uncover the logical flaws of LLMs, problematic cases will be attributed to five error types from two dimensions, i.e., \emph{evidence selection process} and \emph{reasoning process}." The example above seems clearly a reasoning process kind of error, as the LLM is very clear that both the hypothesis and the premise are `identical statements'. But from that it concludes that the hypothesis is \textbf{not} entailed by the premise.

\subsection{Contradictions must be symmetric}
Most humans would agree that if a sentence $A$ is contradictory with a sentence $B$, then sentence $B$ is contradictory with $A$. That is, being contradictory is a symmetric property.  Work in \cite{kalouli-etal-2017-correcting} showed that the humans annotating the corpus SICK did not realize when they had non-symmetric contradictions. We hence checked whether LLMs evaluated contradictions symmetrically. This small experiment showed that out of 495 pairs (5 times 93 contradiction pairs), 49 contradictions were not symmetric. This is not as bad as humans did in the paper above, but it still shows a lack of consistency.

\subsection{Entailment requires premises and hypothesis}
The premise-only work in NLI points to the fact that 
the labels E, C, and N could be accurately determined without any premise, simply using the hypothesis. To make sure that our corpus does not have the same problem, we run an experiment using a dummy true premise,
say, ``Right adjoints preserve limits".  


We substitute this sentence for the premise in all 279 pairs, and evaluate the new pairs using the Group 2 Models. 
These models do not suffer from the same problems that earlier ones did; all four 
essentially classified all of the hypotheses as N, which  is correct.

\begin{table}[h]
    \centering
    \begin{tabular}{llll}\toprule
    Model & E & N & C  \\ \midrule
    L3.1 & .039 & .961 & .0 \\
    Q2 & .004 & .992 & .004 \\
    Mixtral & .0 & 1.00 & .0 \\
    Ge2 & .004 & .996 & .004\\ \midrule
    \end{tabular}
    \caption{Result of Hypothesis only Baseline}
    \label{tab:model:bias:result}
\end{table}

\section{Final remarks}

%
We find it useful to discuss our work by seeing how it aligns with the perceptive conclusions drawn by  \cite{madaan2024lostInInference}.\footnote{We would compare with other sources, but  \cite{madaan2024lostInInference} seems to be the most relevant contemporary paper on this topic.}
We agree that evaluating models on NLI tasks is still relevant.  For Math NLI, we do not find models to be saturated.  This contrasts with ordinary language NLI (ONLI).
We also confirm their finding that ``while the similarity of model distributions with human label distributions increases with
scale, it is still much higher than the similarity between two populations of humans, making it
a potentially interesting statistic to consider.''   We have found that models show less of a distribution of labels than humans.
Finally, they note a certain ``subjectivity'':  ``examples with ‘incorrect’ predictions are rarely in fact incorrect; most
concern questions on which humans may disagree as well.''   And just as they point out,
``The ground truth labels for NLP benchmarks are often decided according to the majority label by human annotators. This simplifies the data annotation process while also making the evaluation easier. However, several previous studies have noted that human disagreements in annotations for NLP datasets reflect the lack of a single ground truth label, rather than noise in the annotation process.''   Even in mathematical texts, there is room for disagreements between experts.

\subsection{Conclusion and future directions}
This paper investigates the performance of Large Language Models (LLMs) on Natural Language Inference (NLI) tasks within the domain of research-level mathematics. We explore the complexities of mathematical language compared to everyday language and evaluate LLMs' ability to handle mathematical inferences, noting some surprising strengths and weaknesses.

 Contrary to what we initially assumed 
 Math NLI is not much easier than ONLI for LLMs. Challenges include unfamiliar vocabulary (e.g., `ring', `field', `comonad'), multimodal elements like symbols and equations, lack of open-source mathematical resources, and the difficulty of finding expert human annotators.

LLMs show paradoxical performance on math tasks: despite exhibiting impressive capabilities in complex mathematical-like problem-solving, LLMs surprisingly struggle with basic logical reasoning and NLI tasks in mathematics. We have documented issues with self-consistency, which is crucial in mathematics. A sanity check testing whether LLMs correctly identify that a statement entails itself ($A \to A$) revealed a very small number of errors, but the explanations for these errors showed a fundamental reasoning flaw.

Post-GPT LLMs avoid some issues that plagued earlier systems.  For example, we expected lexical ambiguity involving math words to cause LLMs to stumble, as in mixing up ``stack'' (a mathematical concept) with ordinary ``stack'' (pile).
They did not do so.

 \rem{LM:We need to say more here.  I moved this from the beginning of the paper, where it was out of place.  It had a pointer to quote \cite{madaan2024lostInInference}.
 And Valeria says:
Well it's trying to say what people have said about NLI that it assumes that the hypothesis is read in the context of the premise, the paper by Madaan deals with the idea that " examples with ‘incorrect’ predictions are rarely in fact incorrect; most concern questions on which humans may disagree as well." also they say that "We find that accuracies (as computed on the majority label) are higher if the entropy of the human labels is low;
when humans disagree, models are more likely to select one of the less preferred labels." I think we need to state some position on this issue, so I believe yes, we need one paragraph here, but not the one that's written, clearly.}

We provide two corpora\footnote{They, along with all of our data, may be found at
\url{https://anonymous.4open.science/r/NLIMath-3FE2}.\label{footnoteOne}}
intended to support further research in the Math NLI area.
One had hypotheses which we wrote ourselves,  and the other had LLMs write the hypotheses.
We believe that these corpora will help newcomers to this attractive area.  And our
results give some idea of what is reasonable to expect from this area in the next years.

Further work directions include combining our work with theorem provers or other symbolic methods,
tests of similarity as opposed to inference, and interactions of our work with running systems in the Math NLI area.

\section*{Limitations}
We did not fine-tune to mathematical text the LLMs we use.  We also only ran things once.
All of our mathematical work was centered on category theory, since that was the source
of our premise pairs.   We do not expect significant differences when we pivot to
other branches of mathematics.

\bibliography{reference}

\appendix
\section{On the LLMs used in this work}
See Table~\ref{tab:10:LLMs}. We used Qwen2-72B-Instruct, which was released in June 2024. According to the Qwen2 Technical Report, this model outperformed Llama3-70B-Instruct on most benchmarks, including mathematical benchmarks such as GSM8K and MATH.

\section{Disagreements between models and humans in the seed corpus}

\begin{enumerate}
 \item 
    \textit{P: Using these ideas, we also prove that magnetic monopoles form an abelian group.} {H: Using these ideas, we also prove that monopoles form an abelian group.}
    
    Humans say the label is N, as it's only for magnetic monopoles that we have the abelian group. Machines say entailment E, but no mathematician would state the weaker result, if they could prove it without the extra hypothesis.
  \item 
    \textit{P: The problem of relating a factorization system to a pointed endofunctor is considered.
} {H: A pointed endofunctor cannot be related to a factorization system.
} 

Humans disagree: some say contradiction C, others say N
  \item 
    \textit{P: This paper introduces the notions of vector field and flow on a general differentiable stack.
} {H: This paper generalizes the notions of vector field and flow on a stack.
} 
  \item 
    \textit{P: We define eventually cyclic Boolean flows and the eventually cyclic spectrum of a Boolean flow.
} {H: The definition of the eventually cyclic spectrum of a Boolean flow uses the definition of eventually cyclic Boolean flows.
} 
  \item 
    \textit{P: The axioms resemble those for monoidal Abelian categories with the addition of an involutive functor.
} {H: The axioms are the ones of monoidal Abelian categories.
} 
  \item 
    \textit{P: The category of Set-valued presheaves on a small category B is a topos.
} {H: The category of Set-valued presheaves on a small category C is a topos.
} 
  \item 
    \textit{P: The category of Set-valued presheaves on a small category B is a topos.
} {H: There exists a small category C such that the category of Set-valued presheaves on C is not a topos.
} 
  \item 
    \textit{P: Various concerns suggest looking for internal co-categories in categories with strong logical structure.
} {H: We suggest looking for internal co-categories.
} 
\end{enumerate}

\section{Seed corpus prompt}
\label{section-seed-corpus-prompt}
Here is the prompt which we used on the seed corpus:

\begin{tcolorbox}[colframe=gray!75!black, boxsep=1mm, left=2mm, right=2mm, top=1mm, bottom=1mm] 
\texttt{[Begin prompt head]}

Suppose you are a logician. Your job is to determine the inference relation between the premise and the hypothesis. There could be three answers: (1) the hypothesis is entailed by the premise; (2) the hypothesis is neutral to the premise; (3) the hypothesis contradicts the premise. Please first tell me your answer and explain why this is  your answer.

\texttt{[End prompt head]}

Premise: \texttt{[Premise]}

Hypothesis: \texttt{[Hypothesis]}
\end{tcolorbox}

\Addresses

\end{document}